\documentclass[letterpaper, 10 pt, conference]{ieeeconf} 
\usepackage{graphicx} 
\title{\LARGE \bf BURNS: Backward Underapproximate \\ Reachability for Neural-Feedback-Loop Systems}
\author{Chelsea Sidrane \and Jana Tumova}
\date{October 2024}
\usepackage{amsmath}
\usepackage{amssymb,amsfonts}

\usepackage{amsthm}
\usepackage{optidef}
\usepackage[linesnumbered]{algorithm2e}

\SetCommentSty{mycommfont}
\usepackage{array}
\usepackage[table,xcdraw,svgnames]{xcolor}
\usepackage{multirow}
\usepackage{url}
\usepackage{cleveref}
\usepackage{ wasysym}
\usepackage{todonotes}
\usepackage{subfig}
\usepackage{lipsum}
\newcommand{\Runder}[2]{\mathcal{R}^{\hat{f}_{cl}}_{(-#1)_{\text{under}}}(\mathcal #2)}
\newcommand{\Runderf}[2]{\mathcal{R}^{f_{cl}}_{(-#1)_{\text{under}}}(\mathcal #2)}
\newcommand{\Rback}[2]{\mathcal{R}^{f_{cl}}_{(-#1)}(\mathcal #2)}
\newcommand{\Rbacknf}[2]{\mathcal{R}^{f_{cl}}_{(-#1)}(#2)}
\newcommand{\Rbackapp}[2]{\mathcal{R}^{\hat{f}_{cl}}_{(-#1)}(\mathcal #2)}
\newcommand{\Rbackappnf}[2]{\mathcal{R}^{\hat{f}_{cl}}_{(-#1)}( #2)}
\newcommand{\bndc}[0]{\subseteq_{\delta c}}

\newtheorem{theorem}{Theorem}[section]
\newtheorem{lemma}[theorem]{Lemma}

\newtheorem{definition}[theorem]{Definition}

\begin{document}

\maketitle
\begin{abstract}
 Learning-enabled planning and control algorithms are increasingly popular, but they often lack rigorous guarantees of performance or safety. 
 We introduce an algorithm for computing underapproximate backward reachable sets of nonlinear discrete time neural feedback loops.
We then use the backward reachable sets to check goal-reaching properties.
Our algorithm is based on overapproximating the system dynamics function to enable computation of underapproximate backward reachable sets through solutions of mixed-integer linear programs.
We rigorously analyze the soundness of our algorithm and demonstrate it on a numerical example. 
 Our work expands the class of properties that can be verified for learning-enabled systems.
\end{abstract}

\section{Introduction \& Related Work}
Neural network control and planning are becoming increasingly prevalent in complex robotic systems~\cite{krinner2024mpcc++, an2024scalable, lee2024learning}. 
Learning-based methods have become popular because they are able to demonstrate superior performance to traditional methods~\cite{song2023reaching}.
However, these methods often lack guarantees of reliability and correctness.
It is difficult to establish such guarantees because neural networks are difficult to analyze -- they are typically non convex and may have anywhere from hundreds to billions of parameters. 

One essential tool for analyzing the reliability and correctness of these systems is known as reachability analysis. 
Reachability analysis involves computing all possible states that the system may reach, and then using those sets to check a \textit{property}~\cite{baier2008principles}.
A property is a system specification written as a logical predicate. 
Two classic properties analyzed in reachability analysis are \textit{reach} properties, which involve reaching a goal set, and \textit{avoid} properties, which involve avoiding an unsafe set.
 Reachability analysis can be performed either forward or backward. 
\textit{Forward} reachability analysis assumes that the system begins within a starting set $\mathcal X_{s}$ and computes subsequent reachable sets into the future. 
\textit{ Backward} reachability analysis begins from some set $\mathcal X_f$ and computes reachable sets into the past that will reach set $\mathcal X_f$~\cite{baier2008principles}.

It is typically intractable to perform reachability analysis exactly for anything more complex than linear or affine systems.
As a result, one usually computes approximations of the reachable set.
Approximate reachability algorithms generally compute either underapproximations,  where the approximate set at time $t$ is a subset of the true reachable set at time $t$, or overapproximations, where each approximate set at time $t$ is a superset of the true reachable set at time $t$. 
Different combinations of underapproximation, overapproximation, forward reachability and backward reachability are useful for checking different kinds of properties.
Overapproximate forward
and backward reachability are both useful for \textit{avoid} properties  because these methods ``inflate'' the unsafe set, capturing all unsafe states and some safe states too. 
Underapproximate backward reachability is useful for \textit{reach} properties
because this method ``shrinks'' the goal set backwards in time,  guaranteeing to capture a subset of the states that will definitely reach the goal.

\begin{figure*}[t]
    \centering
    \includegraphics[trim={0cm 3.5cm 0cm 3.5cm},clip,width=.85\textwidth]{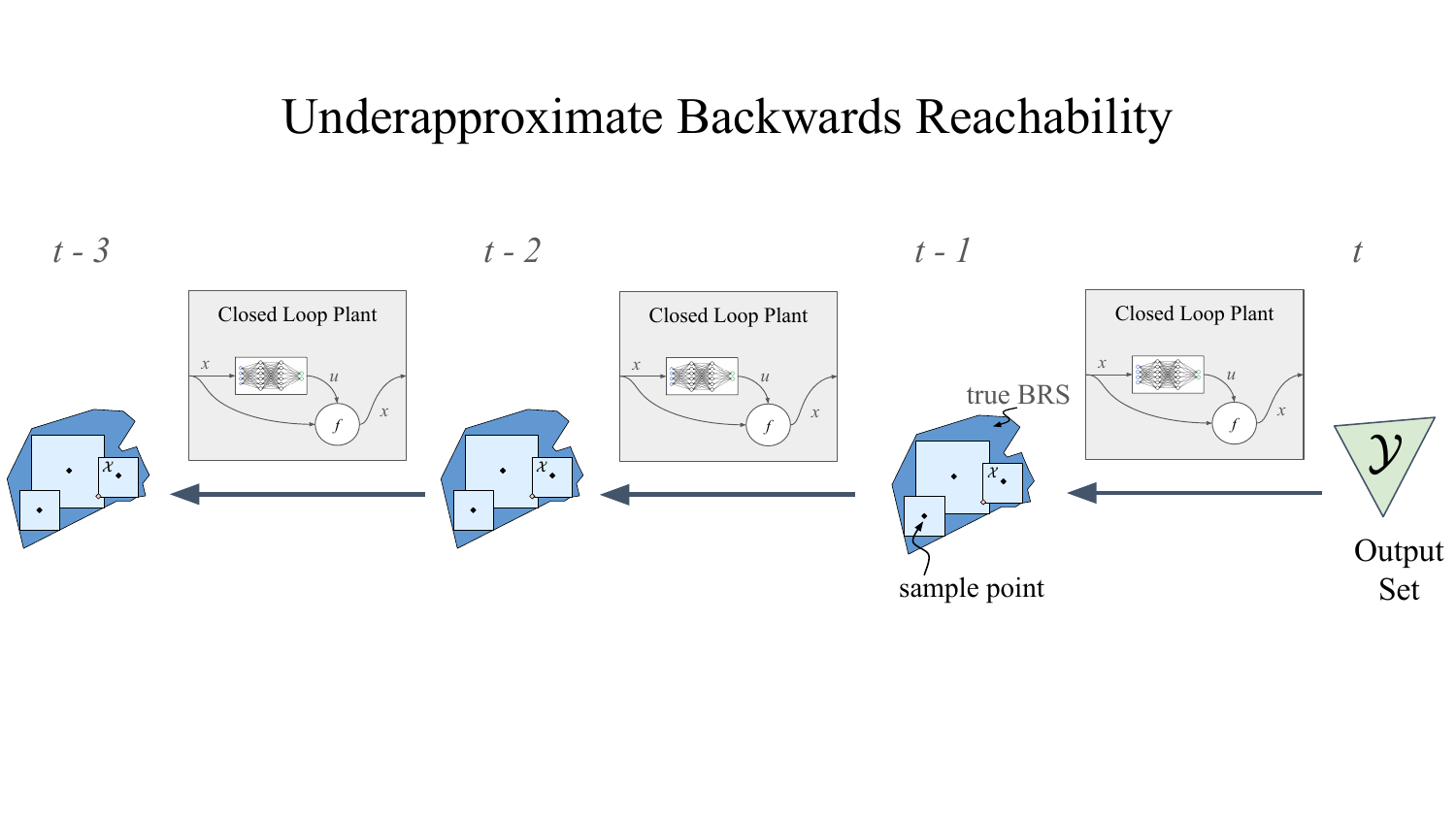}
    \caption{Visual illustration of algorithm}
    \label{fig:illustration}
\end{figure*}

\subsection{Related Work}
There is extensive work on reachability analysis for dynamical systems/transition systems without neural networks in both continuous and discrete time. 
For autonomous systems, some of this work focuses on linear systems~\cite{wetzlinger2023backward}, some on piecewise affine systems~\cite{kloetzer2006reachability} and other work on general nonlinear systems~\cite{chen2013flow}.
Reachability for autonomous nonlinear systems is still an active area of research even without the added complexity of neural network controllers~\cite{abate2024arch}.
Further, there is also extensive work on the Hamilton-Jacobi (HJ) reachability setting wherein one computes the reachable set as well as a control policy~\cite{bansal2017hamilton}. 
However, in this work we focus on reachability for autonomous systems where the controller is a neural network, which requires specialized tools. 

Work on reachability analysis for autonomous systems with neural network control policies, \textit{neural feedback loops}, is a new and growing research area.
There is work on exact reachability analysis for NN dynamics~\cite{vincent2021reachable}, on overapproximate forward~\cite{everett2021reachability, sidrane2022overt} 
and overapproximate backward~\cite{rober2023backward} reachability analysis.
There is even underapproximate backward reachability for linear systems~\cite{everett2021reachability}, but to the best of the authors' knowledge, there is no work on underapproximate backward reachability analysis for general nonlinear neural feedback loops.
As aforementioned, underapproximate backward reachability is the type of analysis needed to ensure that the system starts within a region that will satisfy a \textit{reach} property. 
For this reason, we focus on addressing this gap in literature.

To this end, we take inspiration from methods that seek to verify neural networks in isolation.
Many neural network verification approaches may be seen as forward reachability problems where one is trying to compute the image of  an input set through the neural network function, and then reason about whether that image satisfies a desired property~\cite{liu2021algorithms}.
Analogously, we propose that one may conceptualize of robustness analysis of neural networks as backwards reachability.
Typically, robustness verification for classification tasks, e.g., mapping an image to a label such as `cat', works by computing a region around a specific training image where the label does not change~\cite{dvijotham2018dual}.
The robust region around the training point is then a set in input space which maps to the label `cat.'
A single robustness region does not capture all images that could be labelled `cat' by the network, meaning that it is an underapproximation of the backward reachable set of the label `cat'.
Recent work has expanded on these ideas to explore new ways to compute preimages of a network~\cite{zhang2024premap}.

This paper takes methodological inspiration from network robustness literature to compute under approximate backward reachable sets of nonlinear neural feedback loops.
Robustness verification methods such as \cite{dvijotham2018dual} work by assuming the complement of the property that one wants to hold, e.g. `not cat', parameterizing a ball in the input space around a point (i.e., image) in the training set, and computing the minimum radius of the ball such that `not cat' still holds --  any smaller and the label is always `cat.' 
One of the core contributions of this work is to extend this concept to multi-timestep backward reachability for nonlinear dynamical systems with neural network components. 

\subsection{Contributions}
 The contributions of this work are:
 \begin{itemize}
     \item An algorithm for underapproximate backward reachability of discrete time nonlinear neural feedback loops
     \item  A rigorous theoretical analysis of the algorithm demonstrating its soundness
     \item  A numerical demonstration of the algorithm
     \item  A method for checking inclusion of a polytope in arbitrary non convex sets formed from unions of polytopes which enables the checking of goal reaching properties
 \end{itemize}

\section{Preliminaries}\label{sec:pnb}
First we define some standard notions.
A \textbf{polytope} is a convex set defined by the intersection of a finite number of halfspaces $\{x \mid Ax \leq b\}$ where $A \in R^{m\times n},~x \in R^n,~b\in R^m$.     
A \textbf{p-norm ball} $Ball_p(c, \epsilon)$ centered at point $c$ of radius $\epsilon$ is a set $\{x\mid||x-c||_p\leq\epsilon\}$ where $x,c \in R^n,~\epsilon \in R$.
The set $\mathcal S_{over}$ \textbf{overapproximates} set $\mathcal S$ if $  \mathcal{S} \subseteq \mathcal{S}_{\text{over}}$.
The set $ \mathcal{S}_{\text{under}}$ \textbf{underapproximates} $\mathcal S$ if $ \mathcal{S}_{\text{under}} \subseteq \mathcal{S}$.
We define \textbf{$k$ repeated applications of a function} $f$ by $f\circ^k(x) = f(f(f(\cdots f(x)))$.

A \textbf{feed-forward neural network} is a function mapping $R^n \rightarrow R^m$ with neurons arranged in $p$ layers where the output $y = NN(x)$ is computed $z_1 = (W_1x+b_1),~ z_i = \sigma_i(W_iz_{i-1} + b_i), i\in 2\ldots p$ and $y = z_p$ with weight matrices $W_i \in R^{d_i\times d_{i-1}}$, biases $b_i \in R^{d_i}$, intermediate neurons $z_i \in R^{d_i}$ and activation functions $\sigma_i \in R\rightarrow R$ applied elementwise.
Pre-activation values  are defined $\hat{z}_i = W_iz_{i-1} + b_i$.


\begin{definition}\label{def:closure}
    The \textbf{closure} of a set $S$ is the union of $S$ and its boundary $\delta S$: $\mathbf{cl}S = S \bigcup \delta S$ or equivalently, the intersection of all closed sets containing $S$.
\end{definition}

\begin{definition} \label{def:int}
    The \textbf{interior} of a set $S$, $\mathbf{int}S$, is the set of all interior points in $S$.
\end{definition}

\begin{definition} \label{def:compl}
    For a set $S$ in topological space $X$, the \textbf{complement} $S^c$ is the set of points in $X$ but not in $S$, i.e., $S^c = X \setminus S$.
\end{definition}

\section{Problem Setting}
Consider a discrete-time \textbf{dynamical system} that is described by a system state $x_t \in R^n$ and a dynamics function $f$ that produces the state at the next step as a function of the current state and  the current control input $x_{t+1} = f(x_t, u_t)$; $u_t \in R^m$.
If the control policy providing the control signal $u_t$ is provided and fixed, $u_t = c()$, we refer to this as an \textbf{autonomous system}.
If the control signal $u_t$ at each time step is generated by a neural network policy $u_t = NN(x_t)$, we call this dynamical system a \textbf{neural feedback loop}.
The \textbf{closed-loop} dynamics function, also known as the plant, is defined $f_{cl}(x_t) = f(x_t, NN(x_t))$.

We are interested in computing backward reachable sets under the closed-loop plant $f_{cl}$.
\begin{definition} \label{def:breach}
The exact $\mathbf{k}$\textbf{-step backward reachable set} of $\mathcal G$ for autonomous system $f_{cl}$ is defined 
\begin{align}  \label{eq:breach_def}
\Rback{k}{G} \triangleq \{ \vec{x}_{-k} \mid \vec{x}_{-k+1} &= f_{cl}(\vec{x}_{-k}), \\ \nonumber
\vec{x}_{-k+2} &= f_{cl}(\vec{x}_{-k+1}),\\ \nonumber
& \ldots, \\ \nonumber
\vec{x}_{0} &= f_{cl}(\vec{x}_{-1}), \\ \nonumber
\vec{x}_0 &\in \mathcal G\} \nonumber
\end{align} 
\end{definition}
Once we have computed backward reachable sets, we are interested in checking goal-reaching properties. 
\begin{definition}\label{def:goalreach}
    We say that a neural feedback loop $f_{cl}$ satisfies a \textbf{goal-reaching specification} with goal $\mathcal G$ and horizon $k$ if $\forall x \in \mathcal X_s. \exists t \in[0\ldots k]. f\circ^t(x) \in \mathcal G$. 
\end{definition}
\begin{definition}\label{def:brgr}
A goal-reaching specification holds for goal set $\mathcal G$ if the starting set of the system is a subset of the finite horizon transitive closure of backward reachable sets: 
\begin{equation} \label{eq:brgr}
\mathcal X_{s} \subseteq \bigcup_{t=1}^k\Rback{t}{G}
\end{equation}
\end{definition}
However, if we assume $f$ is an arbitrary nonlinear function, it is not possible to compute exact backward reachable sets.
We must therefore approximate, and our choice of approximation is guided by our intention to check goal-reaching properties.
\begin{lemma}
Computing underapproximate backward reachable sets $\Runderf{t}{G} \subseteq \Rback{t}{G}$ allows for the sound verification of goal-reaching properties.
\end{lemma}
\begin{proof}
If $\mathcal X_{s} \subseteq \bigcup_{t=1}^k\Runderf{t}{G}$ and $\Runderf{t}{G} \subseteq \Rback{t}{G}$ this implies $\mathcal X_{s} \subseteq \bigcup_{t=1}^k\Rback{t}{G}$.
\end{proof}
To summarize, we state the problem as follows. 
\subsection{Problem Statement} \label{sec:prob_state}
Given a nonlinear neural feedback loop (NFL) $x_{t+1} = f(x_t, NN(x_t))$ where $x_t \in R^n$, $u_t \in R^m$ and a goal set $\mathcal G \subseteq R^n$, compute a sequence of $k$ underapproximate backward reachable sets $\Runderf{t}{G}$ such that $\forall t\in1\ldots k.\Runderf{t}{G} \subseteq \Rback{t}{G}$. 



\section{Method}\label{sec:bgnd}
Our approach to the problem defined in \cref{sec:prob_state} is 
to encode the closed-loop plant into a mixed integer linear program (MILP) and solve a series of MILPs over time horizon $k$.
We first explain how one may encode a nonlinear neural feedback loop into the constraints of an MILP,  and we then explain the high level algorithm for computing underapproximate backward reachable sets as well as the specifics of the optimization problems that are solved at each time step.
 We follow the exposition of the algorithm with a rigorous theoretical analysis of its soundness.

\subsection{Encoding a Nonlinear NFL into an MILP}\label{sec:encoding}
Encoding the nonlinear dynamics function $f$ and neural network control policy $NN$ require special consideration to keep the optimization problem in the class of mixed integer linear problems. 
If encoded naively, the problem could become a nonlinear optimization for which we cannot be certain of obtaining the optimal solution. 



Previous literature has developed overapproximate forward reachability algorithms for nonlinear neural feedback loops~\cite{sidrane2022verifying} that involves overapproximation of the dynamics function $f$. 
In a nutshell, the dynamics function is abstracted through re-writing and construction of piecewise linear upper and lower bounds $\forall x \in \mathcal D .f_{i_{LB}}(x) \leq f_i(x) \leq f_{i_{UB}}(x)$ for each nonlinear function $f_i$ 
resulting from rewriting.

We use the same overapproximation technique to treat nonlinear dynamics but construct the optimization problem such that we are able to obtain underapproximations of the backward reachable set (see \cref{lem:oa1} and \cref{lem:oa_many}).
In particular, we define the resulting abstraction $\hat{f}$ as a multi-valued piecewise-linear function that overapproximates the dynamics function $f$.
We define multi-valued function and function overapproximation as follows:
\begin{definition} \label{def:multi}
    As opposed to a single-valued function, a \textbf{multi-valued function} is a function $g$ mapping $\mathcal X \subseteq R^d \rightarrow \mathcal Y \subseteq R^p$ where for each element $x \in \mathcal X$, multiple values of $y\in \mathcal Y$ may belong to the mapping, e.g., $(x_1,y_1) \in g$, $(x_1,y_2) \in g$, and $(x_1,y_3) \in g$.
\end{definition}
\begin{definition}\label{def:funapp}
    An \textbf{overapproximation} $\hat{f}$ of a function $f$ where both map $x \in R^d \rightarrow y \in R^p$ is such that $\forall x,y.(x,y)\in f \implies (x,y)\in \hat{f}$.
    As a corollary, the image of $\hat{f}$ over domain $\mathcal X \subseteq R^d$ overapproximates the image of $f$ over $\mathcal X$: $Im(f(\mathcal X)) \subseteq Im(\hat{f}(\mathcal X))$.
\end{definition}

After abstraction of the dynamics function $f(x,u)$ to $\hat{f}(x,u)$, which may be represented with piecewise-linear functions, any remaining smooth nonlinearity comes from  the control policy $u = NN(x)$. 
In this work, we limit the activation functions $\sigma_i$ in the neural network to piecewise linear activations such as ReLU, $z_i=\max(\hat{z}_i, 0)$.
As a result, the closed loop system $\hat{f}_{cl} = \hat{f}(x,NN(x))$ may be represented as a multi-valued piecewise-linear function.

As further detailed in \cref{sec:meth}, we compute underapproximate backward reachable sets through the solution of optimization problems. 
Piecewise-linear functions may be represented in an optimization problem using mixed-integer constraints, meaning the closed-loop system $\hat{f}_{cl}$ maybe encoded into the constraints of a mixed-integer linear program (MILP).
To encode ReLU functions $t = ReLU(x)$, such as those present in the control policy, one may use the following constraints, introduced by \cite{tjeng2017evaluating}: 
$
    t \geq 0 \bigwedge
    t \geq x \bigwedge
    t \leq u \delta \bigwedge
    t \leq x - l(1-\delta) \bigwedge
    \delta \in \{0,1\}
$
where $[l,u]$ are bounds on $x$, and $\delta$ is a binary variable.
After abstracting $f$ to $\hat{f}$ using OVERT~\cite{sidrane2022verifying}, $\hat{f}$ contains functions $t = \max_{i=1}^m(x_i)$ and $t = \min_{i=1}^m(x_i)$. 
A $\max(x_i)$ function (and $-\min(-x_i)$) may be encoded using the following constraints, also introduced by~\cite{tjeng2017evaluating}:
$
    x_i \leq t \leq x_i + (u_{\max_i} - l_i)(1 - \delta_i) \bigwedge
    \delta_1 + \cdots + \delta_m = 1 \bigwedge
    \delta \in \{0,1\}^m
$
where each $x_i$ has bounds $[l_i, u_i]$, $u_{\max_i} = \max_{j\neq i} u_j$, and the maximum is only taken over inputs where $u_i \geq l_{\max} =  \max_i l_i$.




\subsection{Backward Reachability Algorithm}\label{sec:meth}
Our approach to computation of underapproximate backward reachable sets for nonlinear NFLs is to perform \textit{symbolic reachability analysis} (as defined in \cite{sidrane2022overt}) through the solution of MILPs.
At each time step, we compute multiple norm balls each underapproximating the backward reachable set.
The non-convex union of these norm balls forms an approximation of the backward reachable set at time step $t$.
The reachability algorithm is illustrated in \Cref{fig:illustration} and described in detail in 
\Cref{alg:basic}.


\RestyleAlgo{ruled}
\begin{algorithm}
\caption{Underapproximate Backward Reachability}\label{alg:basic}
\KwData{$f$, $NN$, $k$, $\mathcal G$, $\mathcal D$}
\KwResult{$\{\mathcal R\}_k$}
 $m \gets$ init\_opt\_problem$()$\;
 $\mathcal O \gets \mathcal G$\; 
 $t \gets 1$\;
 \While{$t<k$}{
  $\mathcal{D}_u \gets$ encode\_control(m, NN, $\mathcal D$)\;
 $o = $ encode\_dynamics(m, $f$, $\mathcal{D}$, $\mathcal{D}_u$)\; 
  \For{$j \in 1:n_{\text{samp}}$}{
  $x_{\text{d}} \gets \text{rejection\_sampling}(\mathcal O, \mathcal D, f, t)$\;
  $c \gets $ encode $x_t \in Ball(\epsilon, x_d)$ \tcp{input constraint}
  encode $o \notin \mathcal O$\;
  $\epsilon^* \gets$ solve\_opt\_prob\_\ref{opt1}()\;
  $\text{Ball}_j \gets Ball(\epsilon^*, x_d)$\;
  delete$(c)$\;
  }
  $\{\mathcal R\}_k[t] \gets \bigcup_j \text{Ball}_j$\; 
  $t ++$\;
 }
\end{algorithm}

In \Cref{alg:basic}, lines 1-3 initialize the procedure, and then the algorithm enters a loop to iteratively compute the backward reachable set at each time step $t$.
First, the neural network and dynamics are encoded in lines 5-6 as described in \cref{sec:encoding}.
Next in lines 7-15, the algorithm solves for a set of $n_{samp}$ norm balls, the union of which underapproximates the backward reachable set.
To compute a single norm ball, we assume it is possible to sample a point $x_d$ from the backward reachable which serves as the center for the norm ball (line 8).
Line 9 encodes that the input at time $-t$ must lie in the norm ball.
Line 10 then encodes the constraint that the state at the final time does not lie within the output set, $x_0 \not \in \mathcal O$, and line 11 solves for the optimal radius $\epsilon^*$, which is used to construct a set in line 12.
Line 13 performs cleanup.
Line 15 stores the union of norm balls computed in lines 7-14. 

The complexity of Alg.~\ref{alg:basic} is dominated by the solution of the MILP (line 11).
Further, the size of the MILP is dominated by the encoding of the dynamics and control policy, rather than input and output constraints. 
If the encoding of the dynamics and control policy (lines 5-6) requires $q$ variables, the algorithm solves $n_{samp}$ MILP problems each containing approximately $t*q$ variables at each timestep~$t$, with the final problem containing $\sim k*q$ variables.
However, the size of the encoding of the dynamics may be adjusted by tuning the tightness of the approximation of $f$.
The solve time of the MILP is also affected by the bounding procedure used during the encoding process of the control policy network, which may be chosen freely. 

Next we elaborate on specific subprocedures.

\subsection{Computation of a Single Norm Ball}
\label{sec:detail} 
\begin{figure}
    \centering
    \includegraphics[trim={3cm 3cm 3cm 0},clip,width=0.8\linewidth]{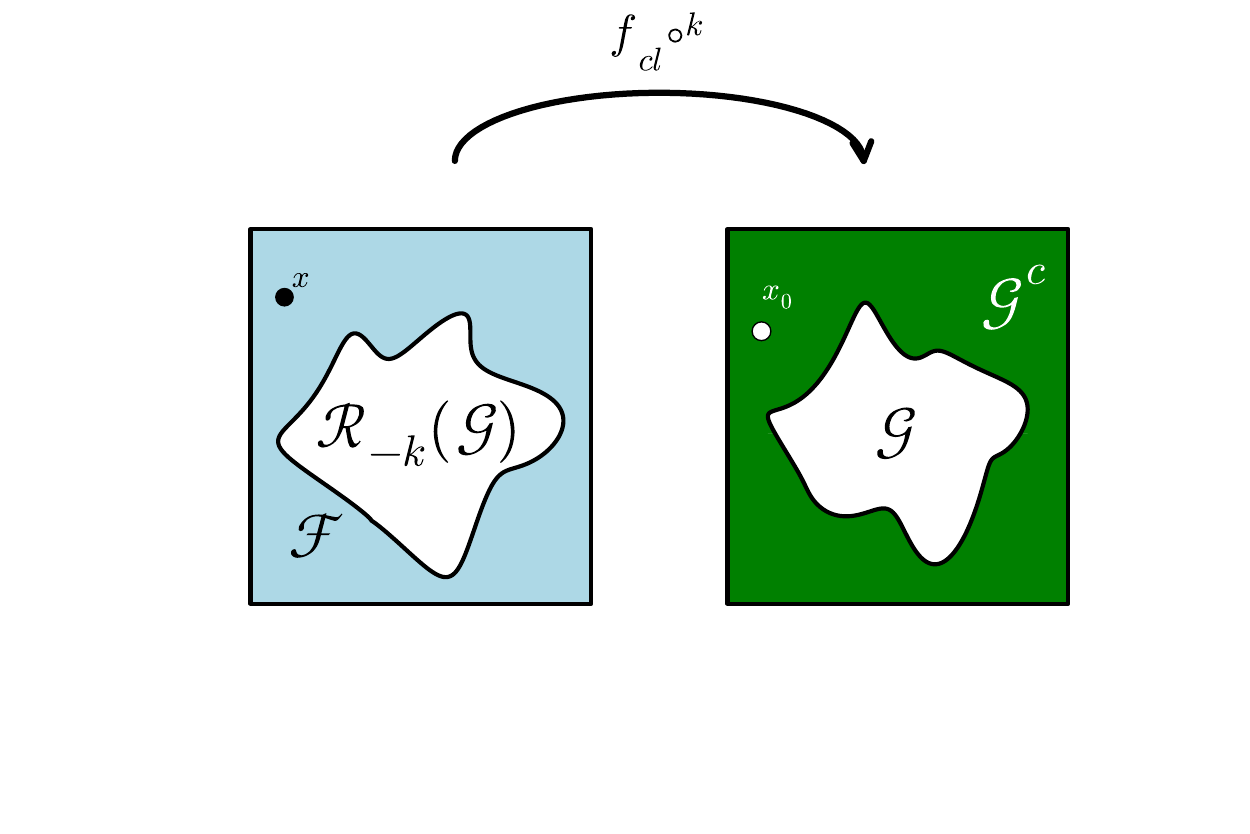}
    \caption{Illustration of sets relevant to proof of \cref{lem:touch}.}
    \label{fig:proof_pic1}
\end{figure}
Wlog, assume that we would like to  compute the  underapproximate backward reachable set of $\mathcal G$, $k$ steps backward: $\Runder{k}{G}$. 
First we provide the optimization problem that we solve to compute a single norm ball within the backward reachable set in \cref{opt1}, and follow with theoretical justification.
Assume that we can sample a point $x_d$ from the set $\Rback{k}{G}$ and then parameterize a $p$-norm ball of radius $\epsilon$ around said point $\{x \mid ||x - x_d||_p \leq \epsilon\}$. 
We enforce the constraint that $x_0$ is generated by repeated application of the closed-loop plant $x_0 = f_{cl}\circ^k(x)$.
We then apply the constraint that $x_0$ does not lie within the goal, $x_0 \not \in \mathbf{int}\mathcal G$, and find the smallest norm ball satisfying the constraints:
  \begin{mini!}|l|
  {x, x_0, \epsilon}{\epsilon}{\label{opt1}}{} 
  \addConstraint{||x-x_{\text{data}}||_p}{\leq \epsilon \label{opt1:con1}}
  \addConstraint{x_0}{=f_{cl}\circ^k(x) \label{opt1:con2}}
  \addConstraint{x_0}{\notin \mathbf{int} \mathcal G \label{opt1:con3}}
 \end{mini!}
 An illustration of relevant sets is shown in \cref{fig:proof_pic1}.
We assume $\mathcal G$ and $\Rback{k}{G}$ as closed sets containing both interior and boundary. 
We also assume that we obtain the optimum solution to opt. prob.~\ref{opt1} and we denote the optimum norm ball as $Ball_p(x_d, \epsilon^*)$.
We claim that $Ball_p(x_d, \epsilon^*)$ is the largest possible $p$-norm ball underapproximation of the backwards reachable set centered at $x_d$.
We argue that this is true because $Ball_p(x_d, \epsilon^*)$ is what we call a boundary coincident subset centered at $x_d$.
If $Ball_p(x_d, \epsilon^*)$ were any larger in radius, it would exceed the backward reachable set and no longer serve as a sound underapproximation. 

\begin{definition}
    \textbf{Boundary Coincident Subset} We call a set $A$ a boundary coincident subset of set $B$, $A \bndc B$, if $A\subseteq B$ and $\delta A \cap \delta B \neq \emptyset$.
\end{definition}
\begin{lemma}\label{lem:closure}
    If $A$ is an open set and $\mathbf{cl}B$ is a closed set, $A \subseteq \mathbf{cl}B \implies \mathbf{cl} A \subseteq \mathbf{cl}B$.   
\end{lemma}
\begin{proof}
    By \cref{def:closure}, $\mathbf{cl}A$ is a subset of any closed set containing A. $\mathbf{cl}B$ is a closed set containing $A$, therefore $\mathbf{cl} A \subseteq \mathbf{cl}B$.
\end{proof}
\begin{lemma}\label{lem:touch}
    $Ball_p(x_d, \epsilon^*) \bndc \Rback{k}{G}$.
\end{lemma}
\begin{proof}
  To demonstrate that \cref{lem:touch} is true, we first show that $Ball_p(x_d, \epsilon^*) \subseteq \Rback{k}{G}$ and then we show that\\ $\delta Ball_p(x_d, \epsilon^*) \bigcap {\delta \Rback{k}{G}\not=\emptyset}$.

    \paragraph{Showing the subset property.}
  First assume 
    $$\exists x. x \in \mathbf{int}Ball_p(x_d,\epsilon^*) ~\bigwedge~
    x\not \in \Rback{k}{G}$$
The proposition $x\not \in \Rback{k}{G}$ may equivalently expressed as $x \in (\Rback{k}{G})^c$ and note that we denote $(\Rback{k}{G})^c$ as $\mathcal F$ in \cref{fig:side-by-side} and \cref{fig:proof_pic1} as it is the interior of the feasible set of the optimization problem. 
    \begin{figure}[h]
    \subfloat[\raggedright Impossible scenario which presents contradiction.]{
        \includegraphics[trim={6cm 1cm 5.5cm 2cm},clip,width=0.45\columnwidth]{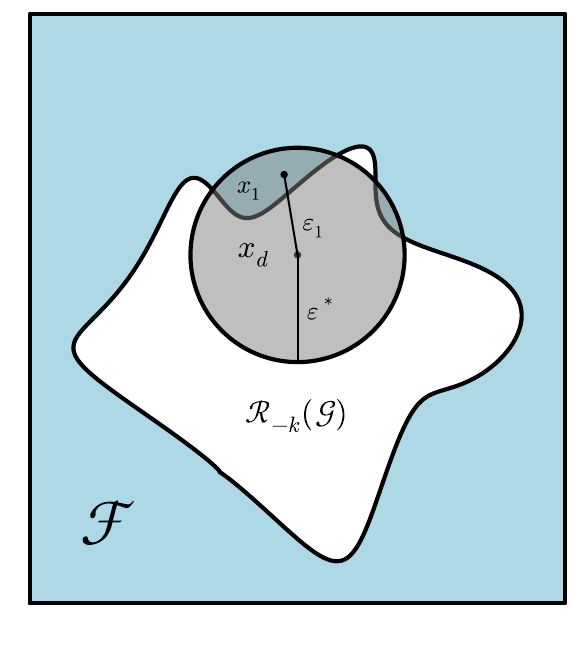}
        \label{fig:impossible}
    }
    \subfloat[Actual outcome of solving \cref{opt1}.]{
        \includegraphics[trim={6cm 1cm 5.5cm 2cm},clip,width=0.45\columnwidth]{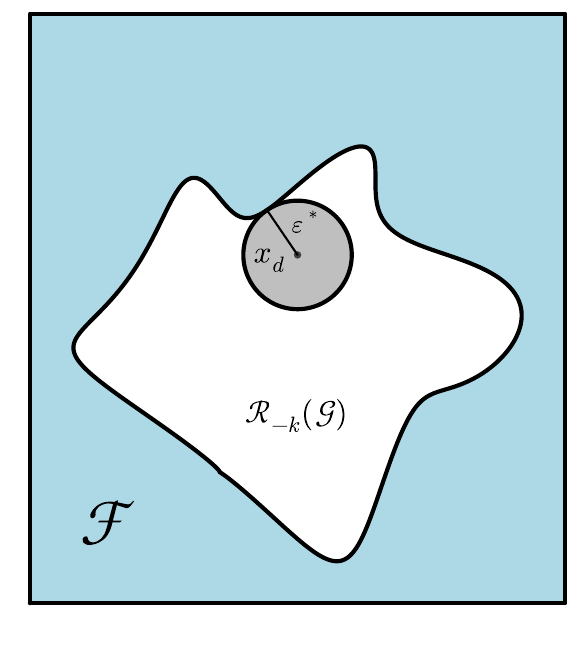}
        \label{fig:actual}
    }
    \caption{Comparison of two figures side by side}
    \label{fig:side-by-side}
\end{figure}

    If there exists a point $\exists x_1.x_1 \in (\Rback{k}{G})^c \bigwedge x_1 \in \mathbf{int}Ball_p(x_d,\epsilon^*)$, we can then define its distance to $x_d$ to be $\epsilon_1 =||x_1 - x_d||_p$.
    It follows that $\epsilon_1 < \epsilon^*$ because all points $\in \mathbf{int}Ball_p(x_d, \epsilon^*)$ will have distance to the center less than the radius $\epsilon^*$.
    Because there exists a point with smaller $\epsilon$, the distance $\epsilon^*$ is therefore not the minimizer of \cref{opt1}, presenting a contradiction; see \cref{fig:impossible} for illustration. 
    Therefore we can assert $\nexists x.x \in (\Rback{k}{G})^c ~\bigwedge~x \in \mathbf{int}Ball_p(x_d,\epsilon^*)$. 
    Note that both sets are open and we have not precluded their boundaries from overlapping.
    It follows that all points in the ball lie inside the backward reachable set $$\forall x.x \in \mathbf{int}Ball_p(x_d,\epsilon^*) \implies x \in \Rback{k}{G}$$ 

    By \cref{lem:closure} and the fact that $\Rback{k}{G}$ is a closed set, we can say that $\mathbf{cl}(\mathbf{int}Ball_p(x_d,\epsilon^*)) \subseteq \Rback{k}{G}$ or rather that 
    $$Ball_p(x_d,\epsilon^*) \subseteq \Rback{k}{G}$$
\paragraph{Showing the Boundary Intersection Property.}
    However, now two scenarios remain, (1) The norm ball is in the interior of the backward reachable set $Ball_p(x_d, \epsilon^*) \subseteq \mathbf{int} \Rback{k}{G}$, or (2)~the norm ball is boundary coincident to the backward reachable set $Ball_p(x_d,\epsilon^*)\bndc\Rback{k}{G}$. We show through contradiction that (2) must be the case.

    Assume (1). In this case, constraint \cref{opt1:con3} that $x_0 \not \in \mathcal G$ could not hold, because $\forall x.x \subseteq \mathbf{int} \Rback{k}{G} \implies x_0 \in \mathcal G$ by 
    \cref{eq:breach_def}. 
    This is a contradiction.  
    Therefore, we can conclude (2) $Ball_p(x_d,\epsilon^*)\bndc\Rback{k}{G}$. See \cref{fig:actual} for an illustration.
\end{proof}
Therefore, we can state that $Ball_p(x_d, \epsilon^*)$ is the largest possible $p$-norm ball underapproximation of the backwards reachable set centered at $x_d$.

\subsection{Generalization to Multiple Underapproximate Sets}
Next, we reason that it is sound to replace $f$ with $\hat{f}$ in the computation of the backward reachable set underapproximation. 
\begin{lemma} \label{lem:oa1}
    If $\hat f: \mathcal X \rightarrow \mathcal Y$ where $\mathcal X \subseteq R^d,~\mathcal Y \subseteq R^k$ and $\mathcal Y = \{\hat f(x) \mid x\in \mathcal X\}$ is a multivalued function 
    that overapproximates the (single-valued) function $f: \mathcal X \rightarrow \mathcal Z$, where $~\mathcal Z \subseteq R^k$, and $\mathcal Z = \{f(x) \mid x\in \mathcal X\}$ 
    the exact 1-step backward reachable set of $\mathcal Z$
    for $\hat{f}$, then $\Rbackapp{1}{Z}$, is an underapproximation of the backward reachable set of $\mathcal Z$ for $f$, $\Rback{1}{Z}$, i.e., $\Rbackapp{1}{Z} \subseteq \Rback{1}{Z}$.
\end{lemma}
    \begin{proof}
    If $\hat{f}$ is a multivalued function overapproximation of $f$, both defined over $\mathcal X$, this means that 
    for every $x \in \mathcal X$, there exists $y$ such that $(x,y) \in f$ and $(x,y) \in \hat f$, and there may exist other $y_i$ such that multiple tuples $(x,y_1), (x,y_2), \ldots$ may belong to $\hat{f}$.
    In other words, $\mathcal Z \subseteq \mathcal Y$. 
    Note that due to the definition of multivalued function overapproximation, $\not \exists x \in \mathcal X.\exists y_1 (x,y_1) \in \hat{f} \wedge \not \exists y_2 (x,y_2)\in f$.

    From \cref{def:breach}, we can say for the overapproximation $\hat f$ that $\Rbackapp{1}{Y} = \mathcal X$. Considering that $\mathcal Z \subseteq \mathcal Y$, we can assert $\Rbackapp{1}{Z} \subseteq \mathcal X$. Again from \cref{def:breach}, we can state for $f$ that $\Rback{1}{Z} = \mathcal X$, and therefore that $\Rbackapp{1}{Z} \subseteq \Rback{1}{Z}$. 
    

    
    \end{proof}
    \begin{figure}
        \centering
        \includegraphics[trim={6cm 3cm 5cm 3cm},clip,width=0.4\linewidth]{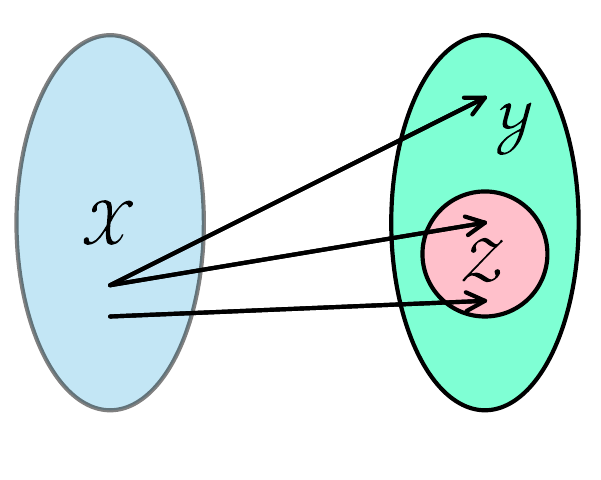}
        \caption{Visual illustration of \cref{lem:oa1}, which demonstrates how we are able to compute underapproximate backward reachable sets from function overapproximations.}
        \label{fig:lem4}
    \end{figure}

Next, we reason about the soundness of using multiple copies of $\hat{f}$ in succession to compute multiple step backward reachable sets: e.g., optimizing $x_t \in Ball_p(\epsilon, x_d),~ \hat{f}\circ^k(x) \in \mathcal G$.
    \begin{lemma}\label{lem:oa_many}
        For each $t$-step, $t\in 1\ldots k$, the backward reachable set of goal set $\mathcal G$ under $\hat f$ underapproximates the $t$-step backward reachable set under $f$: $\forall t\in 1\ldots k.\Rbackapp{t}{G} \subseteq \Rback{t}{G}$.
    \end{lemma}
    \begin{proof}
        We show this by induction. 
        For the base case, corresponding to the first reachable set backward from goal set $\mathcal G$, consider the setting of \cref{lem:oa1}, but now take $\mathcal Z = \mathcal G$ and $\mathcal X = \Rback{1}{G}$.
        We can then assert that $\Rbackapp{1}{G} \subseteq \Rback{1}{G}$.\\
        For the inductive case, we assume that we have two sets such that for some arbitrary $t$, the following holds: 
        \begin{equation}\label{eq:ass}
        \Rbackapp{t}{G} \subseteq \Rback{t}{G}
        \end{equation}
        We then show that  $\Rbackapp{t-1}{G} \subseteq \Rback{t-1}{G}$.  
        The backward reachable set $\Rback{t-1}{G}$ may equivalently be written $\Rbacknf{1}{\Rback{t}{G}}$.
        Invoking lemma~\ref{lem:oa1}, we can state that the one-step backward reachable set of $\Rback{t}{G}$ under $\hat f$ is a subset of that under $f$:
        \begin{equation}\label{eq:almost}
        \Rbackappnf{1}{\Rback{t}{G}} \subseteq \Rbacknf{1}{\Rback{t}{G}}    
        \end{equation}
        And then invoking our assumption in \cref{eq:ass}, we can state
        \begin{equation}\label{eq:there}
        \Rbackappnf{1}{\Rbackapp{t}{G}} \subseteq  \Rbackappnf{1}{\Rback{t}{G}}
        \end{equation}
        Putting \cref{eq:almost} and \cref{eq:there} together, we can then state that 
        $ \Rbackappnf{1}{\Rbackapp{t}{G}} \subseteq \Rbacknf{1}{\Rback{t}{G}}$
        and then rewriting using typical convention, that 
        $$\Rbackapp{t-1}{G} \subseteq \Rback{t-1}{G}$$
        Thus the induction has been shown and we can state that  $$\forall t\in 1\ldots k.\Rbackapp{t}{G} \subseteq \Rback{t}{G}$$
    \end{proof}
And in conclusion we can state:
\begin{theorem}
    \Cref{alg:basic} produces a valid underapproximation of the $k$-step backward reachable set of a set $\mathcal G$.
\end{theorem}
\begin{proof}
If each norm ball computed using opt. prob.~\ref{opt1} is a valid underapproximation per \cref{lem:touch} and \cref{lem:oa_many}, their union is also a valid underapproximation.
Therefore, the union of norm balls produced by \cref{alg:basic} forms a valid underapproxiation of the backward reachable set.
\end{proof}

\subsection{Input and Output Constraints}
 In this section, we  derive how the constraints of the optimization problem can be expressed, as they are represented at somewhat of an abstract level in \cref{opt1}.
The method by which to encode the control policy $u_t = c(x_t)$  and the dynamics $x_{t+1} = f(x_t)$ into the optimization problem \cref{opt1} is described in \cref{sec:encoding}. 
Of particular interest are how the input and output constraints are encoded.
In order to express the output constraint \cref{opt1:con3}, we next present a general method for representing a non-convex union of convex sets using mixed-integer constraints. 

\subsubsection{Encoding a Union of Convex Sets}\label{sec:NONCONVEX}
Consider a convex polytope $\mathcal S = \{ x \mid Ax \leq b \}$, $x\in R^n, A\in R^{m\times n}, b\in R^m$.
We would like to represent the constraint $x \not \in \mathbf{int} \mathcal S$ where $\mathbf{int}\mathcal S = \{x \mid Ax < b\}$.
This may be equivalently written 
$x \in (\mathbf{int}\mathcal S)^c$
where 
\begin{equation}\label{def:setcompl}
(\mathbf{int} \mathcal S)^c = \{x \mid \bigvee_i a_ix \geq b_i\}
\end{equation}
and where $a_i$ is the $i^{th}$ row of $A$ and $b_i$ is the $i^{th}$ element of $b$.
We use binary variables to represent the disjunction of linear constraints. 
Specifically, we re-write \cref{def:setcompl} as
\begin{equation}\label{eq:maxexpr}
\{ x \mid \max(a_1x - b_1, a_2x - b_2, \ldots, a_mx - b_m) \geq 0 \}
\end{equation}
We can then use the encoding for the $\max$ operator described in \ref{sec:encoding} to encode \cref{eq:maxexpr} into a mixed-integer linear program.
As a note on complexity, the number of binary variables needed to represent the constraint $x \in (\mathbf{int}\mathcal S)^c$ grows linearly in $m$, the number of halfspaces defining $\mathcal S$.

If we assume that goal set $\mathcal G$ in optimization problem~\ref{opt1} is a polytope, we may then encode the output constraint~\ref{opt1:con3}, $x_0 \notin \mathbf{int} \mathcal G $, into the MILP using \cref{eq:maxexpr}. 


\subsubsection{Input Constraints}
Input constraint~\ref{opt1:con1}, $||x-x_{d}||_p\leq \epsilon$, defines a cone constraint.
If $p=1$ or $p=\infty$, the resulting expression may be encoded using linear constraints, e.g., 
$||x||_1 \leq t$ may be encoded $-z_i\leq x_i \leq z_i~,~\sum_iz_i\leq t$ and $||x||_{\infty} \leq t$ may be encoded $-t \leq x_i \leq t$.
If $p=2$ is chosen, the problem would become mixed integer convex, as the 2-norm is a convex function and $||x-x_{\text{data}}||_2\leq \epsilon$ would constrain the problem to sublevel sets of a convex function.
Note one is still able to obtain the global optimum for mixed integer convex problems (as for MILP).

To encode the input constraint, we assume that it is possible to sample a point $x_d \in \Rback{t}{G}$.
We achieve this through rejection sampling from  a Sobel sequence~\cite{lavalle2006planning} over a predefined domain $\mathcal D$.
The algorithm is shown in \cref{alg:samp}.

\RestyleAlgo{ruled}
\begin{algorithm}
\caption{Rejection Sampling}\label{alg:samp}
\KwData{$\mathcal O$, $\mathcal D$, $f$, $t$}
\KwResult{$x_d$}
 \While{$true$}{
    $x_d \gets sample(\text{global\_sobel\_sampler}, \mathcal D)$\;
    $y \gets f\circ ^t(x_d)$\;
    \If{$y \in \mathcal O$}{
        return $x_d$\;
    }
 }
\end{algorithm}

The set $\mathcal{D}$ is defined heuristically given simulation traces and adjusted if it is determined to not fully contain the backward reachable sets.

\subsection{Checking Goal Reaching Properties}\label{sec:goalcheck}
In order to check goal reaching properties, we need to check if the starting set of states of the dynamical system, $\mathcal X_s$, is a subset of the finite horizon transitive closure of backward reachable sets, as defined in \cref{def:goalreach} and \cref{def:brgr}: $\mathcal X_{s} \subseteq \bigcup_{t=1}^k\Rbackapp{t}{G}$.
This problem reduces to checking if convex set $\mathcal X_s$ is a subset of a non-convex set represented as a union of convex polytopes.
We introduce an optimization-based approach to test this subset property.

The property $P \subseteq Q$ may equivalently expressed $\forall x.x \in P \implies x \in Q$ which is also logically equivalent to 
\begin{equation}\label{eq:mathbasis}
\not \exists x. x \in P \wedge x \not \in Q
\end{equation}
\Cref{eq:mathbasis} may be expressed as an optimization problem, where we search for the existence of a point $x$ that satisfies the formula.
If no such point exists, we can state $P\subseteq Q$.

If we define the starting set as a polytope, $\mathcal X_s = \{Cx \leq d\}$ with $C \in R^{m \times n}$, $d \in R^n$, the first constraint $x \in \mathcal X_s$ may be easily expressed.
Next, the formula $x \not \in  \bigcup_{t=1}^k\Rbackapp{t}{G}$ may be encoded into the optimization problem using a generalization of the procedure in \cref{sec:NONCONVEX}.
The procedure in \cref{sec:NONCONVEX} allows us to represent the constraint $x \not  \in Q$ where $Q$ is convex. 
Now, we have the setting that $x \not \in Q$ where $Q$ is non-convex because the finite horizon transitive closure of backward reachable sets is a union of a union of convex sets:
$\bigcup_{t=1}^k\Rbackapp{t}{G} = \bigcup_{t=1}^k \bigcup_{j=1}^{nsamp} Ball_j^{(t)}$.
To express that $x \not \in \bigcup_{t,i} Ball_i^{(t)}$, we can equivalently write $x \not \in Ball_1^{(1)} \wedge x \not \in Ball_{nsamp}^{(1)} \wedge \ldots \wedge x \not \in Ball_{nsamp}^{(n)}$ or equivalently, $x \in (Ball_1^{(1)})^c \wedge x \in (Ball_{nsamp}^{(1)})^c \wedge \ldots \wedge x \in (Ball_{nsamp}^{(k)})^c$.
Each constraint $x \in (Ball_j^{(t)})^c$ may be encoded using the procedure in \cref{sec:NONCONVEX}.
Thus the optimization problem may be written
\begin{mini}|l|
    {x}{0}{}{}
    \addConstraint{Cx}{\leq d}{}
    \addConstraint{x \in}{\left(Ball_j^{(t)}\right)^c}{\quad j=1\ldots n_{samp}, t=1\ldots k}  \label{eq:opti}
   \end{mini}
where each set $\left(Ball_j^{(t)}\right)^c$ is represented by $2n$ halfspaces for $p=1$ or $p=\infty$.
Note that precisely speaking, using $\geq$ constraints actually enforces $P \subset Q$ rather than the desired $P \subseteq Q$ but a stricter guarantee is acceptable.  
The resulting optimization problem is then a mixed-integer linear program with $2n\times k \times n_{samp}$ binary variables, where $n$ is the state dimension and $k$ the number of timesteps. 

\section{Numerical Example}\label{sec:exp}
We demonstrate the underapproximate backward reachability algorithm on a nonlinear 2-D robot navigation problem from literature~\cite{rober2023backward}. 
The state is the position $\vec{x}_t = [x_t, y_t]^T$ and
the dynamics function $f$ is given by 
\begin{align}
    {x}_{t+1} =& v \cos(\theta_t) \\
    y_{t+1} =& v \sin(\theta_t)
\end{align}
where heading angle $\theta_t = NN(\vec{x}_t)$ is given by a neural network with three layers of 10 neurons each and ReLU activations, $v$ is a constant and sampling time is 1s.
A hyperrectangular goal set of $[x_0, y_0]^T \in [4,6]\times[6,7]$ was used to compute underapproximate backward reachable sets for 7 timesteps.
We then checked the potential starting sets $\mathcal X_{s_1} = [-1.5, -1.] \times [4.1, 4.6]$ and $\mathcal X_{s_2} = [-2.25, -1.75] \times [3.5, 4]$. 

\begin{figure}[t]
    \centering
    \includegraphics[width=\linewidth]{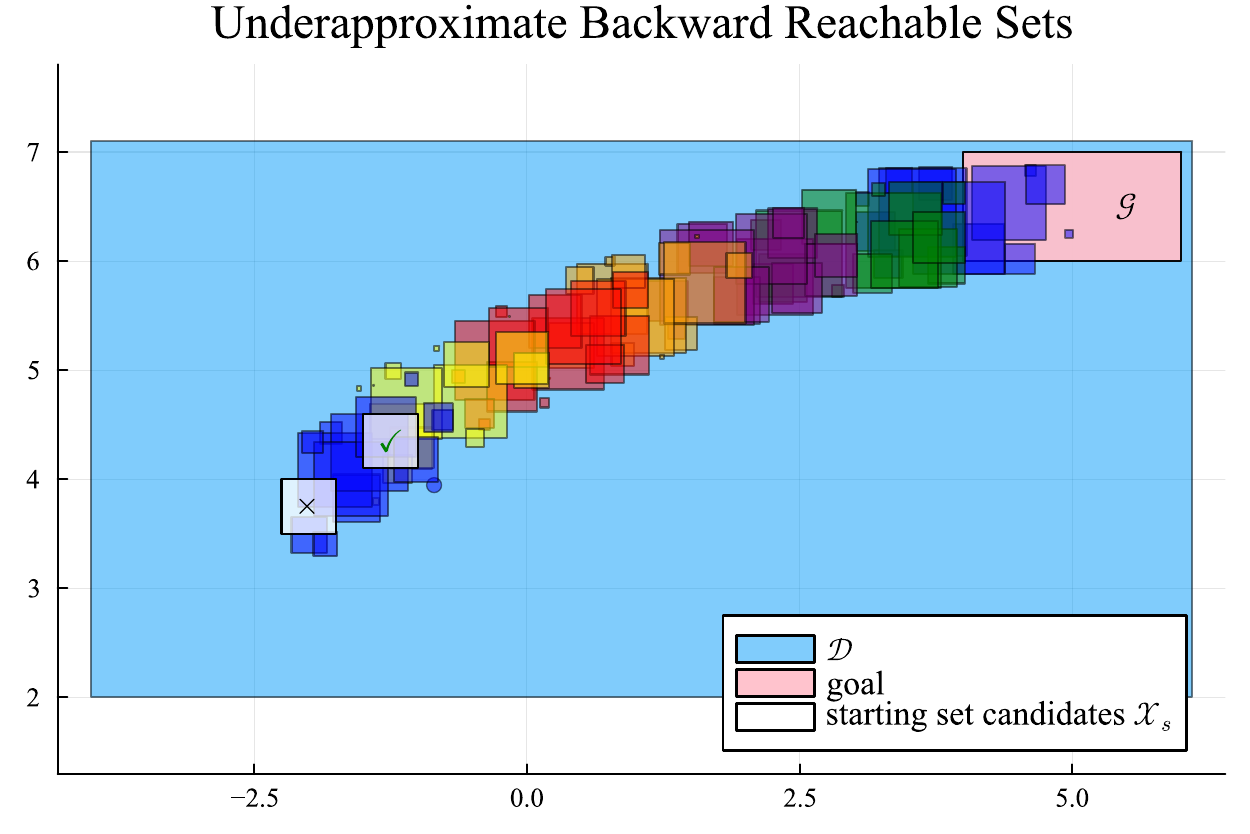}
    \caption{Underapproximate backward reachable sets for seven steps and $n_{samp}=15$, showing two possible starting sets, one safe and one unsafe.}
    \label{fig:1step1}
\end{figure}
\begin{figure}[t]
    \centering
    \includegraphics[width=\linewidth]{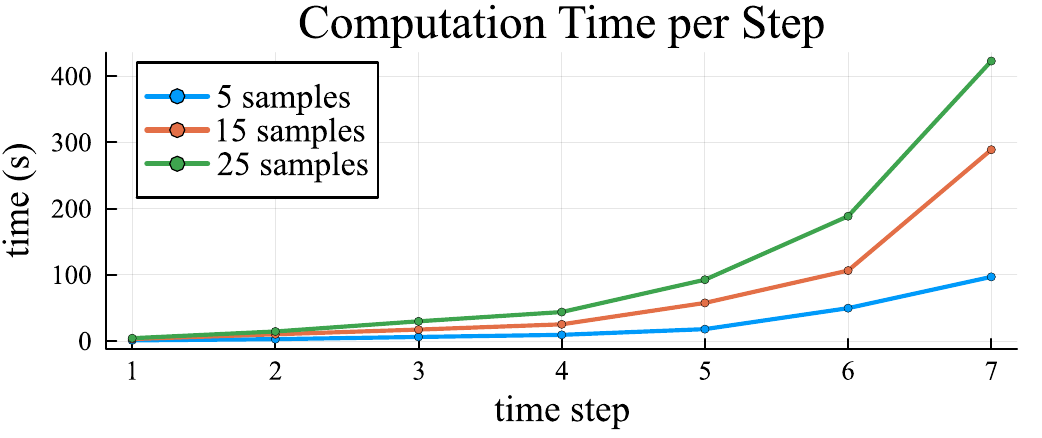}
    \caption{Computation time taken per steps of backwards reachability.}
    \label{fig:time}
\end{figure}

We use the bounding algorithm introduced by~\cite{singh2018fast} to bound the neural network during encoding.
We use the norm $p=\infty$, and a relative error tolerance of $1e^{-6}$ to encode $f$. 
We run an ablation on number of samples with $n_{samp} = \{5, 15, 25\}$.
The reachable sets for $n_{samp}=15$ are shown in \Cref{fig:1step1} and computation time for all are shown in \Cref{fig:time}.
We do not include error bars on the time because the Sobel sampling is deterministic and other sources of randomness are negligible. 
The total time to compute all reachable sets for each number of samples per timestep was 3.1, 8.5, and 13.2 minutes respectively.
Both candidate safe sets shown were checked in $<0.1$ seconds for all settings of $n_{samp}$ using the approach from \cref{sec:goalcheck}, showing the first set satisfies goal reaching for $n_{samp}=[15,25]$ and the second does not for any value of $n_{samp}$.

\begin{figure}
    \centering
    \includegraphics[width=0.6\linewidth]{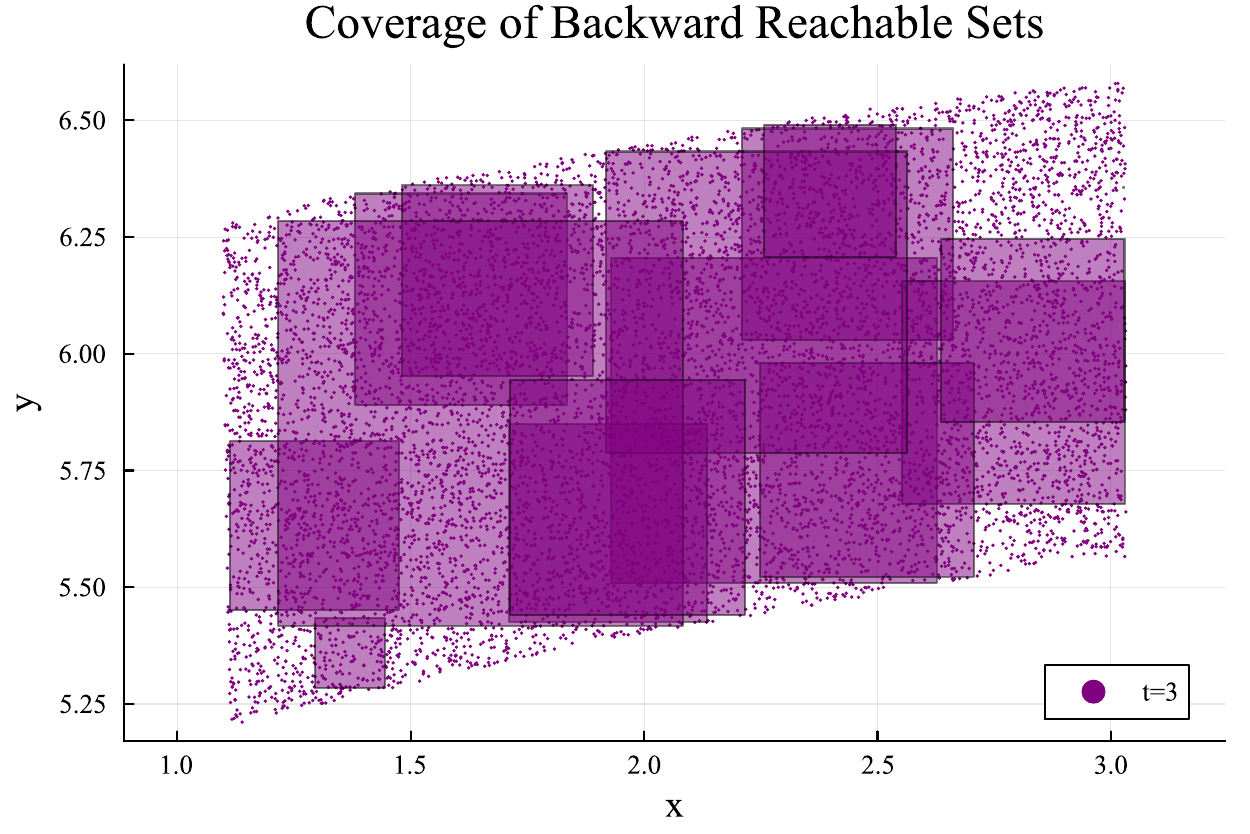}
    \caption{Visualization of underapproximation error of backward reachable sets for $n_{samp}=15$}
    \label{fig:ua_err}
\end{figure}

We measure the underapproximation error of the backward reachable sets using volume fraction of the underapproximate reachable sets as compared to the true reachable set.
We estimate this error using 10,000 uniform random samples per timestep and display the results in \cref{tab:UA}.
We also compute the underapproximation error using all samples and a union of sets over all timesteps, which is in the ``Union'' column of the table. 
We visualize the $81\%$ coverage of the reachable set for $t=3$, $n_{samp}=15$ in \Cref{fig:ua_err}.

\begin{table}[]
\caption{Underapproximation Error Estimates}
\begin{tabular}{llllllllll}
                                                &                            & \multicolumn{7}{c}{\cellcolor[HTML]{EFEFEF}Volume Fraction At Each Timestep}                                                                                                                      &                                                              \\ \cline{3-9}
                                                &                            & \cellcolor[HTML]{EFEFEF}1 & \cellcolor[HTML]{EFEFEF}2 & \cellcolor[HTML]{EFEFEF}3 & \cellcolor[HTML]{EFEFEF}4 & \cellcolor[HTML]{EFEFEF}5 & \cellcolor[HTML]{EFEFEF}6 & \cellcolor[HTML]{EFEFEF}7 & \cellcolor[HTML]{EFEFEF}Union \\
\cellcolor[HTML]{EFEFEF}                        & \cellcolor[HTML]{EFEFEF}5  & 0.34                      & 0.15                      & 0.67                      & 0.23                      & 0.32                      & 0.28                      & 0.34                      & 0.59                                                     \\
\cellcolor[HTML]{EFEFEF}                        & \cellcolor[HTML]{EFEFEF}15 & 0.85                      & 0.89                      & 0.81                      & 0.75                      & 0.84                      & 0.68                      & 0.64                      & 0.88                                                    \\
\parbox[t]{2mm}{\multirow{-3}{*}{\rotatebox[origin=c]{90}{\cellcolor[HTML]{EFEFEF}$n_{samp}$}}}  & \cellcolor[HTML]{EFEFEF}25 & 0.96                      & 0.93                      & 0.90                      & 0.81                      & 0.85                      & 0.80                      & 0.80                      & 0.94                                                     
\end{tabular} \label{tab:UA}
\end{table}

\section{Discussion and Conclusion}\label{sec:conc}
To the best of the author's knowledge, this paper presents the first algorithm for computing underapproximate backward reachable sets of nonlinear discrete-time neural feedback loops.
The algorithm allows one to check goal reaching properties for a class of learning enabled systems that was not previously possible.
The soundness of the algorithm is rigorously analyzed in \Cref{sec:bgnd}, and
the numerical example presented in \Cref{sec:exp} demonstrates that our proposed algorithm is computationally feasible for offline analysis. 

The main limitation of the methodology is scalability; similarly to other verification procedures. 
The algorithm can only analyze a limited finite horizon of steps before the MILP becomes too large to solve. 
\Cref{fig:time} demonstrates how quickly the solve time grows, with the final 7th step taking around $>50\%$ of the total solve time for all values of $n_{samp}$.
Future work to address this limitation is to implement a \textit{hybrid-symbolic} approach as described in \cite{sidrane2022overt, sidrane2024ttt}, which allows for analysis over longer time horizons.

The reasonable underapproximation error for $n_{samp}=[15,25]$ shown in \cref{tab:UA} demonstrates that a limited number of samples can cover most of the backward reachable set. 
We note that underapproximation error increases with timestep which is likely due to accumulated error from the approximation $\hat{f}$.
Also note that $n_{samp}$ is a parameter that can be tuned depending on a user's desire for low error versus speed of computation.

Lastly, a more complete analysis of the algorithm over a variety of example problems would also provide more insight into its performance.
Overall, our work advances the state of the art in verification of learning-enabled systems and lays theoretical groundwork for future algorithms.

\bibliography{my}

\begin{thebibliography}{10}
\providecommand{\url}[1]{#1}
\csname url@samestyle\endcsname
\providecommand{\newblock}{\relax}
\providecommand{\bibinfo}[2]{#2}
\providecommand{\BIBentrySTDinterwordspacing}{\spaceskip=0pt\relax}
\providecommand{\BIBentryALTinterwordstretchfactor}{4}
\providecommand{\BIBentryALTinterwordspacing}{\spaceskip=\fontdimen2\font plus
\BIBentryALTinterwordstretchfactor\fontdimen3\font minus \fontdimen4\font\relax}
\providecommand{\BIBforeignlanguage}[2]{{%
\expandafter\ifx\csname l@#1\endcsname\relax
\typeout{** WARNING: IEEEtran.bst: No hyphenation pattern has been}%
\typeout{** loaded for the language `#1'. Using the pattern for}%
\typeout{** the default language instead.}%
\else
\language=\csname l@#1\endcsname
\fi
#2}}
\providecommand{\BIBdecl}{\relax}
\BIBdecl

\bibitem{krinner2024mpcc++}
M.~Krinner, A.~Romero, L.~Bauersfeld, M.~Zeilinger, A.~Carron, and D.~Scaramuzza, ``Mpcc++: Model predictive contouring control for time-optimal flight with safety constraints,'' \emph{arXiv preprint arXiv:2403.17551}, 2024.

\bibitem{an2024scalable}
T.~An, J.~Lee, M.~Bjelonic, F.~De~Vincenti, and M.~Hutter, ``Scalable multi-robot cooperation for multi-goal tasks using reinforcement learning,'' \emph{IEEE Robotics and Automation Letters}, 2024.

\bibitem{lee2024learning}
J.~Lee, M.~Bjelonic, A.~Reske, L.~Wellhausen, T.~Miki, and M.~Hutter, ``Learning robust autonomous navigation and locomotion for wheeled-legged robots,'' \emph{Science Robotics}, vol.~9, no.~89, p. eadi9641, 2024.

\bibitem{song2023reaching}
Y.~Song, A.~Romero, M.~M{\"u}ller, V.~Koltun, and D.~Scaramuzza, ``Reaching the limit in autonomous racing: Optimal control versus reinforcement learning,'' \emph{Science Robotics}, vol.~8, no.~82, p. eadg1462, 2023.

\bibitem{baier2008principles}
C.~Baier and J.-P. Katoen, \emph{Principles of model checking}.\hskip 1em plus 0.5em minus 0.4em\relax MIT press, 2008.

\bibitem{wetzlinger2023backward}
M.~Wetzlinger and M.~Althoff, ``Backward reachability analysis of perturbed continuous-time linear systems using set propagation,'' \emph{arXiv preprint arXiv:2310.19083}, 2023.

\bibitem{kloetzer2006reachability}
M.~Kloetzer and C.~Belta, ``Reachability analysis of multi-affine systems,'' in \emph{International Workshop on Hybrid Systems: Computation and Control}.\hskip 1em plus 0.5em minus 0.4em\relax Springer, 2006, pp. 348--362.

\bibitem{chen2013flow}
X.~Chen, E.~{\'A}brah{\'a}m, and S.~Sankaranarayanan, ``Flow*: An analyzer for non-linear hybrid systems,'' in \emph{Computer Aided Verification: 25th International Conference, CAV 2013, Saint Petersburg, Russia, July 13-19, 2013. Proceedings 25}.\hskip 1em plus 0.5em minus 0.4em\relax Springer, 2013, pp. 258--263.

\bibitem{abate2024arch}
A.~Abate, M.~Althoff, L.~Bu, G.~Ernst, G.~Frehse, L.~Geretti, T.~T. Johnson, C.~Menghi, S.~Mitsch, S.~Schupp \emph{et~al.}, ``The arch-comp friendly verification competition for continuous and hybrid systems,'' in \emph{International TOOLympics Challenge}.\hskip 1em plus 0.5em minus 0.4em\relax Springer, 2024, pp. 1--37.

\bibitem{bansal2017hamilton}
S.~Bansal, M.~Chen, S.~Herbert, and C.~J. Tomlin, ``Hamilton-jacobi reachability: A brief overview and recent advances,'' in \emph{2017 IEEE 56th Annual Conference on Decision and Control (CDC)}.\hskip 1em plus 0.5em minus 0.4em\relax IEEE, 2017, pp. 2242--2253.

\bibitem{vincent2021reachable}
J.~A. Vincent and M.~Schwager, ``Reachable polyhedral marching (rpm): A safety verification algorithm for robotic systems with deep neural network components,'' in \emph{2021 IEEE International Conference on Robotics and Automation (ICRA)}.\hskip 1em plus 0.5em minus 0.4em\relax IEEE, 2021, pp. 9029--9035.

\bibitem{everett2021reachability}
M.~Everett, G.~Habibi, C.~Sun, and J.~P. How, ``Reachability analysis of neural feedback loops,'' \emph{IEEE Access}, vol.~9, pp. 163\,938--163\,953, 2021.

\bibitem{sidrane2022overt}
C.~Sidrane, A.~Maleki, A.~Irfan, and M.~J. Kochenderfer, ``Overt: An algorithm for safety verification of neural network control policies for nonlinear systems,'' \emph{Journal of Machine Learning Research}, vol.~23, no. 117, pp. 1--45, 2022.

\bibitem{rober2023backward}
N.~Rober, S.~M. Katz, C.~Sidrane, E.~Yel, M.~Everett, M.~J. Kochenderfer, and J.~P. How, ``Backward reachability analysis of neural feedback loops: Techniques for linear and nonlinear systems,'' \emph{IEEE Open Journal of Control Systems}, vol.~2, pp. 108--124, 2023.

\bibitem{liu2021algorithms}
C.~Liu, T.~Arnon, C.~Lazarus, C.~Strong, C.~Barrett, M.~J. Kochenderfer \emph{et~al.}, ``Algorithms for verifying deep neural networks,'' \emph{Foundations and Trends{\textregistered} in Optimization}, vol.~4, no. 3-4, pp. 244--404, 2021.

\bibitem{dvijotham2018dual}
K.~Dvijotham, R.~Stanforth, S.~Gowal, T.~A. Mann, and P.~Kohli, ``A dual approach to scalable verification of deep networks.'' in \emph{UAI}, vol.~1, no.~2, 2018, p.~3.

\bibitem{zhang2024premap}
X.~Zhang, B.~Wang, M.~Kwiatkowska, and H.~Zhang, ``Premap: A unifying preimage approximation framework for neural networks,'' \emph{arXiv preprint arXiv:2408.09262}, 2024.

\bibitem{sidrane2022verifying}
C.~Sidrane, S.~Katz, A.~Corso, and M.~J. Kochenderfer, ``Verifying inverse model neural networks,'' \emph{arXiv preprint arXiv:2202.02429}, 2022.

\bibitem{tjeng2017evaluating}
V.~Tjeng, K.~Xiao, and R.~Tedrake, ``Evaluating robustness of neural networks with mixed integer programming,'' \emph{arXiv preprint arXiv:1711.07356}, 2017.

\bibitem{lavalle2006planning}
S.~M. LaValle, \emph{Planning algorithms}.\hskip 1em plus 0.5em minus 0.4em\relax Cambridge university press, 2006.

\bibitem{singh2018fast}
G.~Singh, T.~Gehr, M.~Mirman, M.~P{\"u}schel, and M.~Vechev, ``Fast and effective robustness certification,'' \emph{Advances in neural information processing systems}, vol.~31, 2018.

\bibitem{sidrane2024ttt}
C.~Sidrane and J.~Tumova, ``Ttt: A temporal refinement heuristic for tenuously tractable discrete time reachability problems,'' \emph{arXiv preprint arXiv:2407.14394}, 2024.

\end{thebibliography}

\end{document}